# Local Connection Reinforcement Learning Method for Efficient Control of Robotic Peg-in-Hole Assembly

Yuhang Gai, Jiwen Zhang, Dan Wu*, and Ken Chen
All authors are from the State Key Laboratory of Tribology in Advanced Equipment,
Department of Mechanical Engineering, Tsinghua University, Beijing, China.
*Corresponding author: Dan Wu (phone: 1-391-083-2965; e-mail: wud@mail.tsinghua.edu.cn).

**Abstract**—Traditional control methods of robotic peg-in-hole assembly rely on complex contact state analysis. Reinforcement learning (RL) is gradually becoming a preferred method of controlling robotic peg-in-hole assembly tasks. However, the training process of RL is quite time-consuming because RL methods are always globally connected, which means all state components are assumed to be the input of policies for all action components, thus increasing action space and state space to be explored. In this paper, we first define continuous space serialized Shapley value (CS3) and construct a connection graph to clarify the correlativity of action components on state components. Then we propose a local connection reinforcement learning (LCRL) method based on the connection graph, which eliminates the influence of irrelevant state components on the selection of action components. The simulation and experiment results demonstrate that the control strategy obtained through LCRL method improves the stability and rapidity of the control process. LCRL method will enhance the data-efficiency and increase the final reward of the training process.

## 1. Introduction

### 1.1. Robotic Assembly Control

Robotic assembly based on off-line planning cannot coordinate stress and guarantee precision between assembly objects [1]. Hence, the robot is usually guided by external feedback of vision or force information to execute assembly tasks. The perception range of vision is limited and it is easy to be affected by the environment, while force information is more intrinsic to feel the stress and pose error between assembly objects. Hence, compliance control methods based on force feedback are more widely used [2].

Compliance control methods used in assembly tasks construct the mapping between force/moment and the relative pose of assembly objects [3]. According to whether the structure and parameters of a controller are adaptive, compliance control methods can be divided into three categories: constant compliance control methods, artificially designed adaptive compliance control methods, and learning-based adaptive compliance control methods [4].

The structure and parameters of a constant compliance controller are pre-configured and always constant in the assembly process [5]-[7]. Constant compliance control methods can only solve simple assembly tasks with weak nonlinear dynamics. However, the dynamics of peg-in-hole assembly tasks are continuously changing and strongly nonlinear. The capacity of the constant compliance controller may be insufficient to handle the peg-in-hole assembly tasks.

Hence, adaptive compliance control methods are proposed to solve assembly tasks to obtain better control performance [8]-[10]. Variable compliance centre and variable compliance parameters are the most common artificially designed adaptive compliance control methods. Variable compliance centre method changes the structure of the controller by converting motion and force/moment information from the robot and sensor to the dynamic compliance centre. Variable compliance parameters method changes the parameters of the controller according to an artificially designed adaptive law. However, the performance of artificially designed adaptive compliance controllers is limited by human experience.

Learning-based adaptive compliance control methods are proposed to obtain better performance on assembly tasks [11]-[14]. RL method is the most widely employed learning method [15]-[21]. RL abstracts an assembly task as a Markov Decision Process (MDP) and supplies a policy to guide each control step of the assembly process. Because learning-based adaptive compliance control methods construct adaptive law through data-driven methods, the adaptive law is more optimal for the current assembly task. Learning-based adaptive controllers perform much better than artificially designed ones in assembly tasks. However, the biggest dilemma is that learning-based methods are time-consuming and not stable enough in industry applications.

### 1.2. Efficient RL-Based Control

RL is widely used in the field of continuous control [22]. Compared with artificially designed controllers, RL tends to use networks to take place of analytic feedback control laws and use exploration and exploitation to take place of empirical designing. However, RL is plagued by the problem of long training sessions [23]-[26], especially in high-dimensional continuous action space and state space. When action space and state space are high-dimensional and continuous, the exploration will become relatively inefficient, thus reducing the data-efficiency and convergence speed of the training process. One key technology to improve data-efficiency and convergence speed is to optimize the dimensionality of action space and state space.

A primary method for accelerating the training process is to construct a decoded mapping from action space or state space to a low-dimensional latent space, and then learn in the latent space [27]-[31]. Because the dimensionality of latent space is smaller, the size of latent space to be explored is reduced, which makes the exploration of RL more efficient. Gaussian Process Regression is usually employed to learn the mapping between latent space and action or state space. In the fields of multi-agent RL and sequence action space RL, extracting and training in latent space is also an effective means to improve data-efficiency and convergence speed [32][33].

Hierarchical RL optimizes dimensionality by decomposing a high-dimensional action space into a sequence of high-level and low-level sub-action spaces [34]-[36]. Then hierarchical RL learns a policy consisting of multiple layers, each responsible for a different level of control. The dimensionality of low-level sub-action spaces is reduced to be smaller than that of original action space. The action space to be explored in hierarchical RL is divided and conquered, thus improving data-efficiency and convergence speed. Compared with training in latent space, hierarchical RL is potential to perform better on a task but suffers from the selection of low-level sub-action spaces.

An action dimensionality extension (ADE) method is proposed in [37], which draws on the ideas of latent space and hierarchical RL but works differently. ADE method first constructs a low-dimensional action space according to the similarity between action components and trains a primitive agent effectively. Then the agent is extended into high-dimensional original space and continues to be trained to obtain a better performance on the task. ADE combines the advantages of higher data-efficiency in low-dimensional action space and better performance in high-dimensional action space.

All the methods mentioned above can accelerate RL by optimizing the dimensionality of action and state space. However, the policies are still globally connected. Specifically, the policy for each action component is decided by all state components. Some irrelevant state components will damage the performance of policies and also decrease the data-efficiency and convergence speed. If the dependence of action components on state components is obtained, the state space could be divided into several sub-state spaces and construct several local connection policies for each action component, which is beneficial to improve data-efficiency and convergence speed.

### 1.3 Motivation and Contribution

Motivated by constructing controllers more efficiently through RL method, this paper first defines CS3 to judge the effect of an action component on a state component in complex tasks. Then a connection graph is constructed based on CS3 to clarify the correlativity of action components with state components and define the input of policies. LCRL method based on the connection graph is proposed to eliminate the influence of irrelevant state components on the policies for action components, thus accelerating the training process.

The main contributions of this paper are listed as follows. First, we propose LCRL method to accelerate convergence and increase the final reward of RL algorithms. LCRL method is based on the definitions of CS3 and the connection graph, which show the dependence of action components on state components. Besides, LCRL method is implemented to construct a learning-based compliance controller of the peg-in-hole assembly task, which causes lower force/moment and guarantees a more stable control process.

The rest of the paper is organized as follows. Section 2 introduces LCRL method. Section 3 develops the control method of robotic peg-in-hole assembly using LCRL method. Sections 4 and 5 provide simulation and experiment verifications. Section 6 summarizes the research work of this paper.

## 2. Local Connection Reinforcement Learning
### 2.1 Reinforcement Learning in Continuous Space

RL abstracts arbitrary control problems into Markov decision processes (MDP) $\mathcal{M}=(\mathcal{S},\mathcal{A},\mathcal{P},\mathcal{R},\gamma)$, where $\mathcal{S}$ is state space, $\mathcal{A}$ is action space, $\mathcal{P}:\mathcal{S}\times\mathcal{A}\times\mathcal{S}\to[0,1]$ is the state transition function, $\mathcal{R}:\mathcal{S}\times\mathcal{A}\times\mathcal{S}\to\mathbb{R}$ is the reward function, and $\gamma\in[0,1)$ is the discount rate. Here we record state space $\mathcal{S}$ as a $m$-dimensional space and action space $\mathcal{A}$ as a $n$-dimensional space. State $s\in\mathcal{S}$ is a $m$-dimensional vector $s=[^1s,^2s,...,^ms]$. Action $a\in\mathcal{A}$ is a $n$-dimensional vector $a=[^1a,^2a,...,^na]$. State transition function is recorded as $p(s_{t+1}|s_t,a_t)$, which gives the probability of next state $s_{t+1}$ once current state $s_t$ and action $a_t$ are determined. State transition function is usually implicit in complex tasks. The targets of tasks are shaped through the reward function $r(s_t, a_t, s_{t+1})$. The long-term value can be evaluated through the sum of discounted rewards.

$$Q_\pi(s,a) = \mathbb{E}_\pi \left[ \sum_{i=0}^{\infty} r(s_{t+i}, a_{t+i}, s_{t+i+1}) \gamma^i \right] \tag{1}$$

The terminal goal of the RL is to train an optimal policy $\pi^*$ to maximize the value function.

$$\pi^*(s) = \arg\max_\pi Q_\pi(s,a) \tag{2}$$

For normalized continuous action space and state space, $\mathcal{S}$ can be taken apart as $m$ orthogonal subspace $^j\mathcal{S}, j = 1, 2, ..., m$. Subspace $^j\mathcal{S}$ involves states, all components of which are zeros except component $^j s$.

$$^j\mathcal{S} = \{s \mid {^k s} = 0, k \neq j\} \tag{3}$$

Similarly, $\mathcal{A}$ can be taken apart as orthogonal subspaces $^i\mathcal{A}, i = 1, 2, ..., n$. Subspace $^i\mathcal{A}$ involves actions, all components of which are zeros except component $^i a$.

$$^i\mathcal{A} = \{a \mid {^k a} = 0, k \neq i\} \tag{4}$$

***note***: $\mathcal{S}$, $\mathcal{A}$, $^j\mathcal{S}$, $^i\mathcal{A}$ are not strictly linear spaces because elements are bounded. Except for the boundedness of elements, $\mathcal{S}$, $\mathcal{A}$, $^j\mathcal{S}$, $^i\mathcal{A}$ meet all the properties of linear space. In the following contents, we still use related concepts of linear space when not causing confusion.

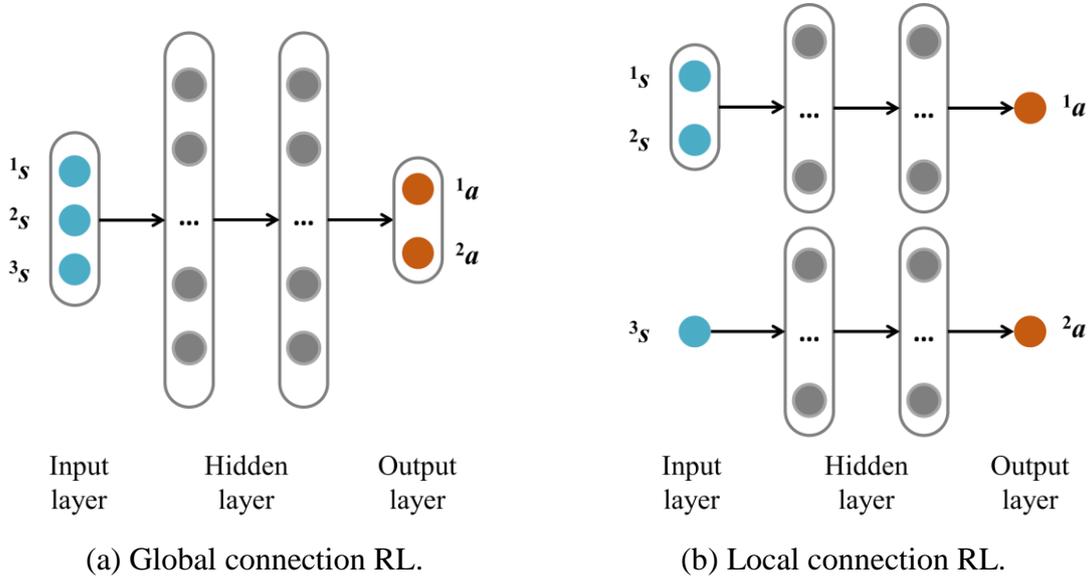

(a) Global connection RL.  (b) Local connection RL.

Fig. 1. Definitions of global connection RL and local connection RL.

At each control step, state $s_t \in \mathcal{S}$ at step $t$ is collected according to the status of the environment. Action $a_t \in \mathcal{A}$ is determined by RL policy $\pi(a_t | s_t)$. As shown in Fig. 1, according to the mapping between action subspaces and state subspaces as well as the dependence of action components on state components, RL methods can be divided into two categories:

***global connection RL*** (**GCRL**): Each action subspace is mapped to all state subspaces and each action component is determined by all state components.

*local connection RL* (**LCRL**): Each action subspace is mapped to part of the state subspaces and each action component is determined by part of state components.

In existing RL methods, policy $\pi(a_t|s_t)$ is globally connected by default. Some irrelevant state components have a negative influence on the selection of action components. Hence, GCRL method could cause potential curse of dimensionality and reduce data-efficiency of the training process. If local connection relation is set appropriately, it is possible to avoid mapping irrelevant state subspaces to some action subspaces, thus ensuring both optimality and data-efficiency. The specific LCRL method is proposed in later subsections.

## 2.2. Connection Graph

The basic idea of LCRL method is that if some action component $^i a$ does not affect some state component $^j s$ during several state transitions, the selection of $^i a$ is independent of $^j s$. The key step is depicting the effect of $^i a$ on $^j s$. As indicated as state transition function, the next state is decided by current state $s_t$ and action $a_t$. $^j s$ may continuously change even action is a zero vector. Key metric of the effect of $^i a$ on $^j s$ is the bias between state trajectories of $^j s$ before and after activating $^i a$. Here we propose a conception of CS3 to judge the effect of $^i a$ on $^j s$.

$$\phi(^i a, ^j s) = \mathbb{E}_{\mathcal{L} \subseteq \mathcal{A} \setminus ^i\mathcal{A}, a^{(1)} \in \mathcal{L}, a^{(2)} \in ^i\mathcal{A}, s_0'=s_0 \in \mathcal{S}} \left( \sum_{t=0}^{m-1} \left| ^j s_{t+1}'(a^{(1)} + a^{(2)}, s_t') - ^j s_{t+1}(a^{(1)}, s_t) \right| \right) \tag{5}$$

where CS3 $\phi()$ describes the effect of $^i a$ to $^j s$ and is in a similar format of Shapley function [38]. The larger CS3 value is, the more $^i a$ affects $^j s$. $\mathbb{E}$ is expectation function. \ is orthogonal complement space function. $\subseteq$ is subspace function. $\mathcal{L}$ represents a subspace of $\mathcal{A} \setminus ^i\mathcal{A}$. $a^{(1)}$ and $a^{(2)}$ are vectors in subspace $\mathcal{L}$ and $^i\mathcal{A}$. In summary, CS3 describes the effect of $^i a$ on $^j s$ through comparing state trajectories before and after adding $a^{(2)}$ in $^i\mathcal{A}$ to $a^{(1)}$ in $\mathcal{L}$.

Since RL-based control methods are always implemented in continuous space, CS3 is defined in the manner of sampling and expectation, which is profitable to decrease algorithm complexity. Besides, another unique feature of CS3 is defined along a state trajectory whose length is the dimensionality of state space $m$ instead of a single state point. That is because an action component may not change a state component in one step and the effect may occur after several steps in some cases. Besides, the integration of $a^{(2)} \in ^i\mathcal{A}$ (set $^i a_t$ to be non-zero) may increase or decrease the coming state $^j s_{t+1}$, causing that the terminal state $^j s_m$ and $^j s_m'$ may be equal. For example, the state trajectory after adding $a^{(2)} \in ^i\mathcal{A}$ is 1→0→1, while the state trajectory before adding $a^{(2)} \in ^i\mathcal{A}$ is 1→1→1. The terminal state are the same but the sum of absolute value of state bias are different, which shows the rationality of CS3. Hence, the serialized feature allows long-term effect to be considered, which is not only beneficial, but also necessary in some cases to reflect the effect of $^i a$ on $^j s$.

Based on CS3 $\phi(^i a, ^j s)$ for arbitrary $i$ and $j$, the effect of all action components on all state components can be defined through a connection graph $G = [G_{ij}] \in \mathbb{R}^{n \times m}$. Each element in $G$ is 1 or 0, which shows whether the action component contributes to changing the state component or not.

$$G_{ij} = \begin{cases} 0, \phi(^i a, ^j s) = 0 \\ 1, \phi(^i a, ^j s) \neq 0 \end{cases} \tag{6}$$

In most application cases, CS3 $\phi(^i a, ^j s)$ always be non-zero even if it is theoretically zero because of sampling noises. Hence, each element $G_{ij}$ is determined through comparing the value of $\phi(^i \mathcal{A}, ^j \mathcal{S})$ and the average value of all $\phi(^k a, ^j s)$, $k = 1, 2, \ldots, n$ in applications.

$$G_{ij} = \begin{cases} 0, \phi(^i a, ^j s) < T \left( \sum_{k=1}^{n} \phi(^k a, ^j s) \right) \Big/ n \\ 1, \phi(^i a, ^j s) \geq T \left( \sum_{k=1}^{n} \phi(^k a, ^j s) \right) \Big/ n \end{cases} \tag{7}$$

where $T$ is the threshold set manually. Empirically, $T$ is usually set to be 0.1 in assembly tasks.

The connection graph clarifies the dependence between all state components and action components. LCRL method is dedicated to remove irrelevant state component inputs from RL agent. The most basic method is to remove state components, which an action component does not affect. Hence, the connection graph exactly defines the structure of the LCRL agent.

The method of building the connection graph is shown in Algorithm 1.

---

**Algorithm 1**: Building connection graph through CS3
**Input**: Threshold $T$, Sampling times $ST$
**Output**: Connection graph $G$
for $i = 1, n$:
  for $st = 1, ST$:
    Random sample $s_0 \in \mathcal{S}$, $\mathcal{L} \subseteq \mathcal{A} \setminus ^i\mathcal{A}$
    for $t = 0, m\text{-}1$:
      Random sample $a^{(1)} \in \mathcal{L}$, $a^{(2)} \in ^i\mathcal{A}$
      for $flag = 0, 1$:
        if $flag = 0$:
          Take action $a^{(1)} + a^{(2)}$ based on $s_t{'}$, observe $s_{t+1}{'}$
          Store $s_{t+1}{'}$ in a buffer $\boldsymbol{B_{c1}}$
        if $flag = 1$:
          Take action $a^{(1)}$ based on $s_t$, observe $s_{t+1}$
          Store $s_{t+1}$ in a buffer $\boldsymbol{B_{c2}}$
  Calculate $\phi(^i a, ^j s)$, $j = 1, 2, \ldots, m$ using (5) according to $\boldsymbol{B_{c1}}$ and $\boldsymbol{B_{c2}}$
Calculate $G$ using (7)

---

## 2.3. LCRL Setup Based on Connection Graph

Based on the connection graph, LCRL provides local connection policies instead of global connection policies. The primary concern is how to design the structure of networks and training mode to obtain local connection policies. As indicated in the connection graph $G$, the elements which are 1 in row $i$ mark state component inputs required by action component $^i a$.

$$^i a = {}^i \pi(^{j_1} s, ^{j_2} s, \ldots), \; ^{j_1} s, ^{j_2} s, \ldots \in \{ ^j s \mid G_{ij} = 1 \} \tag{8}$$

where $j_1$, $j_2$ represent column labels of elements which are 1. $^i \pi$ represents the policy of action component

$^ia$. $^{j_1}s$, $^{j_2}s$,... are the input of policy $^i\pi$.

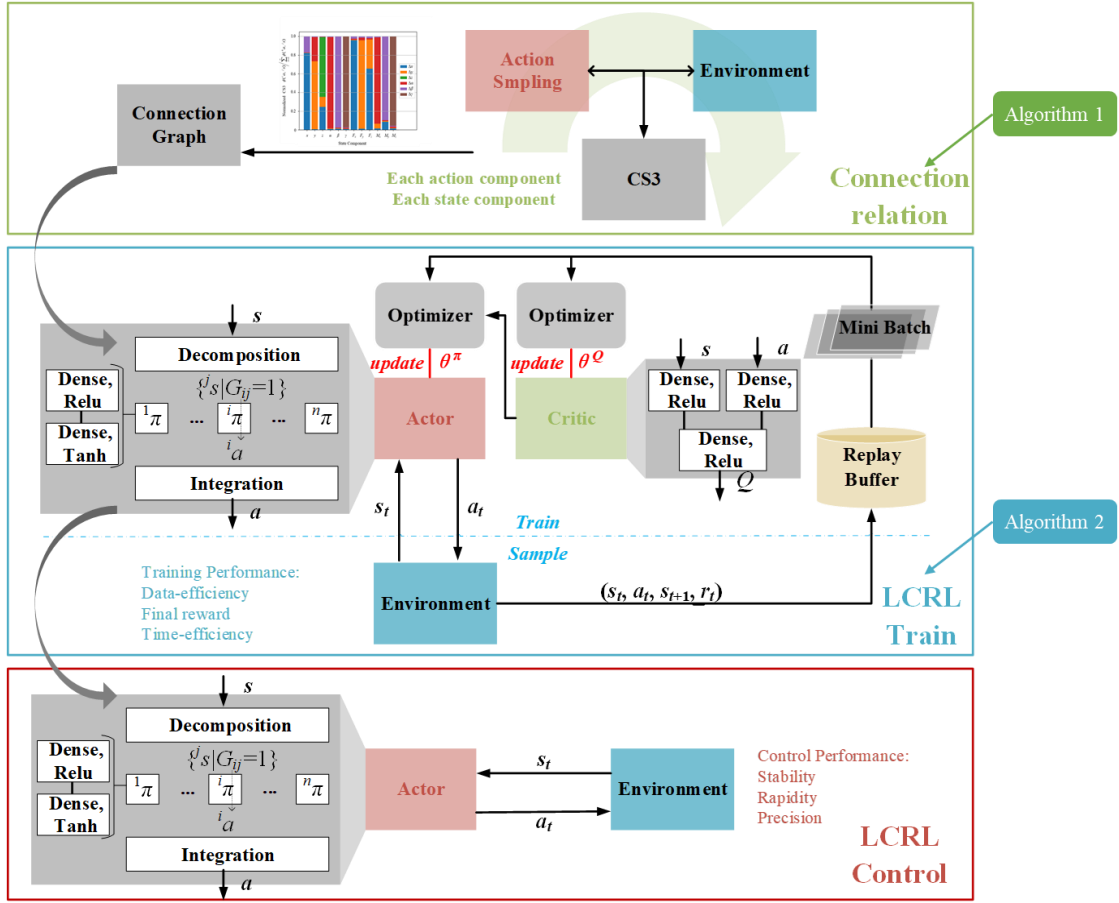

Fig. 2. Flow chart of LCRL method implemented in DDPG algorithm.

**Algorithm 2**: DDPG with LCRL method
**Input**: Connection graph $G$
**Output**: RL agent
Initialize networks
Initialize replay buffer $B$
for $ep = 1$, $Max\_episode$:
   Receive the initial state
   for $t = 0$, $Max\_step$:
     for $i = 1$, $n$:
       Decompose state $s_t$ as $\{^js|G_{ij} = 1\}$
       Select action component $^ia_t$ using policy $^i\pi$
     Integrate action components as action $a_t$
     Execute action $a_t$, observe reward and state $s_{t+1}$
   Store a transition in $B$
   Update networks using minibatch sampled from $B$

As shown in Fig. 2, we choose Deep Deterministic Policy Gradient (DDPG), a typical actor-critic RL algorithm, as a demonstration case for applying LCRL method. The actor is an online policy network used to provide action for current state. The critic is an online value network to evaluate the value of the action-state pair. The structure of LCRL agent is different from those of common GCRL agent. In the actor with LCRL method, multiple subnetworks, a decomposition module, and an integration module are employed to replace the globally connected network. Subnetworks provide policies for each action component. Decomposition module in actor decomposes a global state vector *s* as groups of corresponding local state components for policies of action components following (8). The integration module in actor integrates local action components as a global action vector *a*. The network structure of critic is kept the same as the global connection policy.

DDPG with LCRL method is shown in Algorithm 2.

## 2.4. Explanation in Linear Time-Invariant System

***lemma* 1**: In a linear time-invariant system $s_{t+1} = Es_t + Fa_t, s_t \in \mathbb{R}^m, a_t \in \mathbb{R}^n, E \in \mathbb{R}^{m \times m}, F \in \mathbb{R}^{m \times n}$, elements in row *i* and column *j* of the connection graph *G* and matrix *H* are both zeros or both non-zeros. Matrix *H* is defined as

$$H = \sum_{t=0}^{m-1} C_t$$
$$C_t = \left|(E^t + E^{t-1} + ... + I)F\right|, t = 0, 1, ..., m-1 \tag{9}$$

Here an example is provided where the connection graph *G* is built and LCRL method is implemented in a linear time-invariant system. The system is defined as

$$E = \begin{bmatrix} -1 & 1 & 0 \\ 0 & 1 & 0 \\ 0 & 0 & 1 \end{bmatrix}, F = \begin{bmatrix} 1 & 0 & 0 \\ 0 & 1 & 0 \\ 0 & 0 & 1 \end{bmatrix} \tag{10}$$

According to *lemma* 1, matrix *H* and the connection graph *G* are

$$H = C_1 + C_2 + C_3 = \begin{bmatrix} 1 & 0 & 0 \\ 0 & 1 & 0 \\ 0 & 0 & 1 \end{bmatrix} + \begin{bmatrix} 0 & 1 & 0 \\ 0 & 2 & 0 \\ 0 & 0 & 2 \end{bmatrix} + \begin{bmatrix} 1 & 1 & 0 \\ 0 & 3 & 0 \\ 0 & 0 & 3 \end{bmatrix} = \begin{bmatrix} 2 & 2 & 0 \\ 0 & 6 & 0 \\ 0 & 0 & 6 \end{bmatrix}$$
$$G = \begin{bmatrix} 1 & 1 & 0 \\ 0 & 1 & 0 \\ 0 & 0 & 1 \end{bmatrix} \tag{11}$$

where $C_1$, $C_2$, $C_3$ describe one-step effect, two-step effect, and three-step effect. *H* describes the total effect in three steps. The results demonstrate that $^1a$ affects $^1s$ and $^2s$, $^2a$ affects $^2s$, $^3a$ affects $^3s$. Hence, only relevant state components are configured as the inputs of corresponding action component. The state feedback control law is

$$a_t = Ks_t, K = \begin{bmatrix} k_1 & k_2 & 0 \\ 0 & k_5 & 0 \\ 0 & 0 & k_9 \end{bmatrix} \tag{12}$$

The poles of the system with the state feedback control law above are $1-k_1, 1-k_5, 1-k_9$, which can be assigned arbitrarily. The results indicate that the performance of the system will not get worse even without applying global feedback (each action component is decided by all state components). Meanwhile, due to the reduction of the number of parameters to be solved in *K*, it is simpler to design the feedback control law. If such a connection graph also exists in a nonlinear or time-variant system, we can also optimize the structure of networks to improve data-efficiency.

If we build state feedback control law against the connection graph, for example,

$$a_t = Ks_t, K = \begin{bmatrix} 0 & k_2 & 0 \\ k_4 & 0 & 0 \\ 0 & 0 & k_9 \end{bmatrix} \quad (13)$$

The poles of the system become $\sqrt{1+k_4(k_2-1)}, -\sqrt{1+k_4(k_2-1)}, 1-k_9$, which cannot be assigned arbitrarily because the sum of $\sqrt{1+k_4(k_2-1)}$ and $-\sqrt{1+k_4(k_2-1)}$ is always 0. The performance of the system may be limited because the feedback law contradicts the connection graph.

## 3. Control Method

### 3.1. Formulation of Robotic Peg-in-Hole Assembly

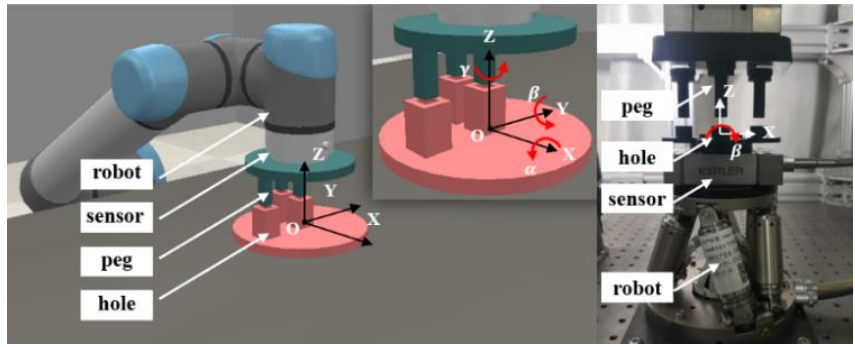

(a) Simulation.　　　　(b) Experiment.
Fig. 3. Robotic peg-in-hole assembly task and manipulation of the robot.

This section focuses on settling robotic assembly tasks using proposed LCRL method. The assembly tasks are accomplished by a robot with a force/moment sensor. As shown in Fig. 3, assembly objects are a peg group and a hole group. The force/moment sensor is installed at the end of the robot. The robot manipulates one of assembly objects to insert the peg into the hole. The force/moment sensor detects force/moment information caused by contact occurring between the peg and hole. LCRL controller continuously coordinates the motion of the robot according to force/moment information to achieve assembly targets. The state is defined as the Cartesian pose of the robot and six-dimensional force/moment detected by the force/moment sensor. State at control step $t$ is

$$s_t = [x, y, z, \alpha, \beta, \gamma, F_x, F_y, F_z, M_x, M_y, M_z] \quad (14)$$

where $x, y, z, \alpha, \beta, \gamma$ are Cartesian pose of the robot, $F_x, F_y, F_z, M_x, M_y, M_z$ are force/moment detected by force/moment sensor.

Action is defined as revision factors of parameters in the compliance controller and influences the translations and rotations of the robot.

$$a_t = [\Delta x, \Delta y, \Delta z, \Delta \alpha, \Delta \beta, \Delta \gamma] \quad (15)$$

All elements of action are set within the range of [$lb$, $ub$], $lb \in \mathbb{R}^6$, $ub \in \mathbb{R}^6$.

The assembly targets are defined through the reward function. The assembly tasks are expected to be accomplished fast and the contact force/moment are expected to be small. Hence, the reward function is defined as

$$r = r_d + r_s \qquad (16)$$

where dense reward $r_d$ is

$$r_d = -h_z z - h_F \|F\|_2 - h_M \|M\|_2 \qquad (17)$$

where $h_z$ is the coefficient for assembly speed and $h_F$, $h_M$ are the coefficients for contact force/moment. $F = [F_x, F_y, F_z]$ and $M = [M_x, M_y, M_z]$ are force and moment detected by the sensor. The first term in reward function is used to encourage the agent to accomplish the assembly task as soon as possible, while the last two terms are used to confine contact force/moment. $r_s$ is the sparse reward to evaluate a training episode. If the peg is inserted to target depth $L$ and forces and moments of the entire assembly process are all within the limitation, the total episode will be rewarded by a positive $r_s$.

### 3.2. Controller Based on LCRL

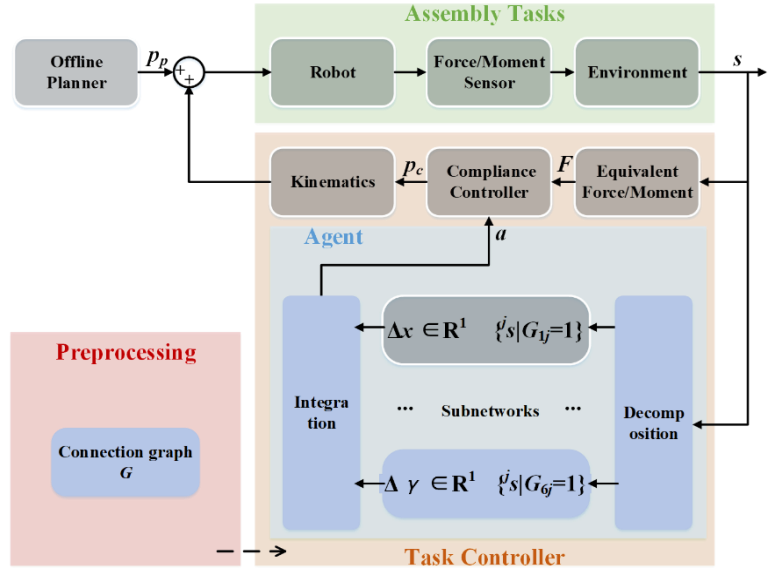

Fig. 4. Adaptive compliance controller combining a compliance controller and an LCRL agent. A connection graph is first constructed in preprocessing and then the LCRL agent is designed according to the connection graph. The action of the LCRL agent revises parameters in the compliance controller.

The controller structure of the assembly task is shown in Fig. 4. An off-line planned trajectory and feedback adjustment drive the robot to execute manipulation together. The off-line planned trajectory is continuously coordinated according to force/moment feedback. The off-line planned trajectory $p_p$ is

$$p_p = [0, 0, L/step, 0, 0, 0] \qquad (18)$$

where $L$ is the target insertion depth. $step$ is the expected control step to accomplish the assembly task.

The control method of the peg-in-hole assembly task is a learning-based adaptive compliance control method. Kinematics, compliance controller, and equivalent force/moment modules follow the format of [8]. Compliance controller is a proportional controller.

$$\hat{p}_c = \hat{K}(F - F_{rfr}) \tag{19}$$

where $F_{rfr}$ indicates reference force/moment. When the pegs and holes are with clearance fit, $F_{rfr}$ = [0, 0, 0, 0, 0, 0]. $F$ indicates the equivalent force/moment after processing force/moment detected by sensor. $\hat{K} = \mathrm{diag}(k_x, k_y, k_z, k_\alpha, k_\beta, k_\gamma)$ is a diagonal matrix of proportional parameters. $\hat{p}_c$ is the output of compliance controller, which will be converted as the pose adjustment to the robot after kinematics module.

The adaptive law is given by RL agent and used to update $\hat{K}$ through

$$p_c = K(F - F_{rfr}), K = \mathrm{diag}(a\hat{K}) + \hat{K} \tag{20}$$

where $a \in \mathbb{R}^6$ is the action of the agent, which represents the adaptive law of compliance parameters. $p_c$ is the output of the adaptive compliance controller, which is composed of a compliance controller and an RL agent. Here a typical learning-based adaptive compliance control framework has been constructed for robotic peg-in-hole assembly tasks.

The RL agent is trained with LCRL method based on the connection graph. As shown in Fig. 4, before training, a connection graph is first obtained following algorithm 1. With the guidance of the connection graph, the agent uses multiple locally connected policies in place of a globally connected policy to reduce the size of networks. The implementation of LCRL method follows algorithm 2. In an LCRL agent, an action integration layer and a state decomposition layer are required to interact with the environment. The decomposition module decomposes state $s$ into corresponding groups of state components $\{^j s | G_{1j} = 1\}$, $\{^j s | G_{2j} = 1\}$, ..., $\{^j s | G_{6j} = 1\}$ for policies of action components. The integration module integrates action components $\Delta x, \Delta y, \Delta z, \Delta \alpha, \Delta \beta, \Delta \gamma$ as an action vector $a$. Specific procedures is shown in subsection 2.3.

## 4. Simulations

### 4.1. Training process

Table I. Parameters in simulations.

| Variable | Value | |
|---|---|---|
| | Group 1 | Group 2 |
| Axis length of pegs | 30 mm | 20 mm |
| Depth of holes | 30 mm | 20 mm |
| Side length of holes | 10 mm | 8 mm |
| Side length of pegs | 9.9 mm | 7.95 mm |
| Target insertion depth | 30 mm | 20 mm |
| **Variable** | **Value** | |
| Threshold $T$ | 0.1 | |
| Target insertion step | 50 | |
| Range of random pose | [± 0.2 mm, ± 0.2 mm, ± 0.2 mm, ± 0.5°, ± 0.5°, ± 0.5°] | |
| Constant compliance | $\hat{K}$ = diag(5e-3, 5e-3, 5e-5, 1e-3, 1e-3, 1e-3) | |
| Action boundaries | lb = [-1, -1, -1, -1, -1, -1], ub = [2, 2, 2, 4, 4, 4] | |
| Reward coefficients | $h_z$=1, $h_F$=0.1, $h_M$=1, $r_s$=2 | |
| Learning rate of DDPG | 1e-3(actor), 1e-2(critic) | |
| Soft update rate of DDPG | 1e-3 | |
| Learning rate of PPO | 1e-4(actor), 1e-3(critic) | |

The main target of the simulation is to show the optimization mechanism of LCRL method. An assembly environment is built in CoppeliaSim software (Fig. 3 (a)). The parameters of assembly objects and controller are listed in Table I. Two groups of assembly objects with different axis lengths, side lengths, and clearances are selected.

In an assembly process, the peg is first moved to align and then inserted into the hole. The alignment pose of the peg is within a random range where the bottom surface of the peg is kept at a small distance above the top surface of the hole. Then an LCRL agent is trained and the adaptive compliance controller is constructed by combining the LCRL agent to guide the insertion process.

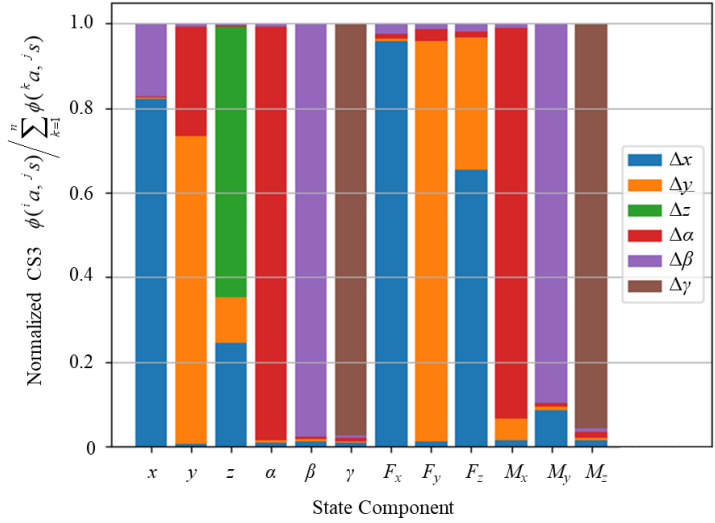

Fig. 5. Normalized CS3 for all action components and state components to construct the connection graph of the peg-in-hole assembly task in simulations.

A connection graph is first constructed according to algorithm 1. For arbitrary $i$ and $j$, normalized CS3 are illustrated in Fig. 5. The connection graph is determined according to (7).

$$G = \begin{bmatrix} G_{11} & G_{12} \\ G_{21} & G_{22} \end{bmatrix}, G_{11} = \begin{bmatrix} 1 & 0 & 1 & 0 & 0 & 0 \\ 0 & 1 & 1 & 0 & 0 & 0 \\ 0 & 0 & 1 & 0 & 0 & 0 \end{bmatrix}, G_{12} = \begin{bmatrix} 1 & 0 & 1 & 0 & 1 & 0 \\ 0 & 1 & 1 & 1 & 0 & 0 \\ 0 & 0 & 0 & 0 & 0 & 0 \end{bmatrix},$$
$$G_{21} = \begin{bmatrix} 0 & 1 & 0 & 1 & 0 & 0 \\ 1 & 0 & 0 & 0 & 1 & 0 \\ 0 & 0 & 0 & 0 & 0 & 1 \end{bmatrix}, G_{22} = \begin{bmatrix} 0 & 1 & 1 & 1 & 0 & 0 \\ 1 & 0 & 1 & 0 & 1 & 0 \\ 0 & 0 & 0 & 0 & 0 & 1 \end{bmatrix}$$
(21)

The connection graph is the same for two groups of assembly objects. As indicated in (21), the peg-in-hole assembly task is divided into two orthogonal planner assembly tasks, which is consistent with intuition. An LCRL agent is configured according to the connection graph and trained to accomplish peg-in-hole assembly tasks. The local connection policy is set as

$$\Delta x = {}^1\pi(x, z, F_x, F_z, M_y)$$
$$\Delta y = {}^2\pi(y, z, F_y, F_z, M_x)$$
$$\Delta z = {}^3\pi(z)$$
$$\Delta \alpha = {}^4\pi(y, \alpha, F_y, F_z, M_x) \quad (22)$$
$$\Delta \beta = {}^5\pi(x, \beta, F_x, F_z, M_y)$$
$$\Delta \gamma = {}^6\pi(\gamma, M_z)$$

Because the LCRL agent needs to be robust to the initial pose error, the initial pose error of the peg is selected randomly within a range in each training episode. The reward curves are shown in Fig. 6. The training processes of GCRL method and LCRL method are repeated several times. Solid curves represent the mean reward trajectories, while the edges of color blocks represent the minimum and maximum reward trajectories. To verify the universality of LCRL method, we implement LCRL method on DDPG (a typical off-policy RL algorithm) and PPO (a typical on-policy RL algorithm) separately.

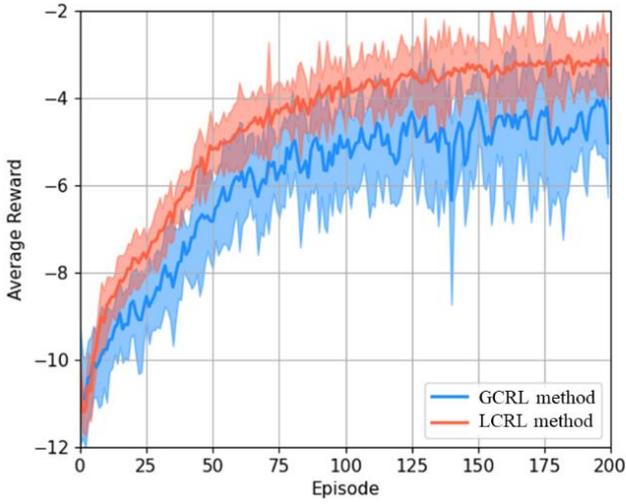

(a) DDPG, assembly objects group 1.

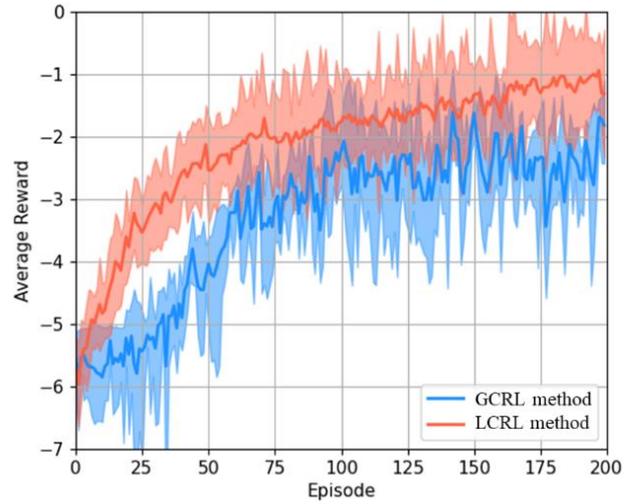

(b) DDPG, assembly objects group 2.

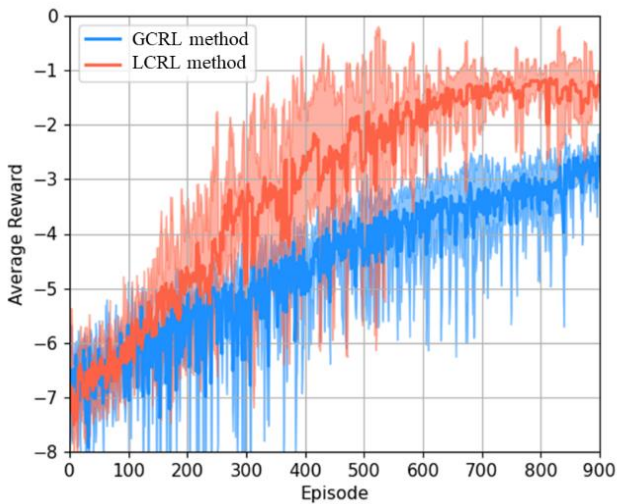

(c) PPO, assembly objects group 1.

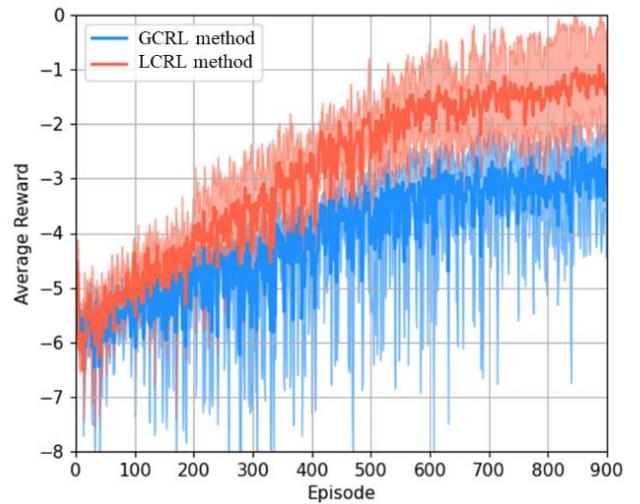

(d) PPO, assembly objects group 2.

Fig. 6. The reward curves of training process in simulations.

Data-efficiency and final reward are the most critical evaluation indicators to characterize the performance of LCRL method. Data-efficiency reflects the convergence speed of the training process of LCRL agent and final reward reflects the performance of LCRL agent on the current assembly task. As shown in Fig. 6, we compare the data-efficiency and final reward of the global connection agent trained directly and the local connection agent trained with LCRL method. The performance of LCRL method on DDPG algorithm is shown in Fig. 6 (a) and (b). The number of episodes for LCRL method to reach -5 average reward is about 50% smaller than that of GCRL method. The final reward of LCRL is about 33% higher than that of GCRL method. The performance of LCRL method on PPO algorithm is shown in Fig. 6 (c) and (d). The number of episodes to reach -3 average reward and final reward are improved about 50% and 37.5% after implementing LCRL method. The results demonstrate that LCRL method is profitable for both data-efficiency and final reward.

Time-efficiency of LCRL method is also verified through comparing the training time before and after implementing LCRL method. The training time of DDPG for 200 episodes and the training time of PPO for 900 episodes are recorded. Only the time spent on optimizing the agent and constructing the connection graph is counted, while the time spent on executing assembly actions is not included. The results demonstrate that training the agent with LCRL method will increase the time cost to a limited extent. The additional time of LCRL method is mainly used to construct the connection graph. Considering the improvement of data-efficiency and final reward, the time spent on establishing the connection graph is indispensable.

Table II. Time-efficiency in simulation training process.

| Agent | Time-Efficiency (s) | | | |
|---|---|---|---|---|
| | GCRL (Total Time) | LCRL (Connection graph) | LCRL (Training process) | LCRL (Total Time) |
| DDPG | 439.1±33.9 | 64.7±0.8 | 410.2±25.9 | 474.9±26.7 |
| PPO | 1109.0±61.7 | 64.3±1.2 | 1087.6±40.3 | 1151.9±41.5 |

**4.2. Testing Process**

In the testing process, the initial pose error of the robot is set to be [0.2 mm, 0.2 mm, 0.2 mm, 0.5°, 0.5°, 0.5°]. The assembly objects are selected as group 1. The RL algorithm is selected as DDPG. We test the performance of three controllers: constant compliance controller, compliance controller combined with GCRL agent, and compliance controller combined with LCRL agent.

In the control process of all three methods, operation forces and moments continue to increase at the beginning of the assembly because the interference area becomes broader. Then operation forces and moments start to approach 0 under the guidance of the compliance controller.

The rapidity and stability of the control process are determined by compliance parameters. A conclusion is that larger compliance parameters render the rapid reduction of force/moment and the suppression of peak values, but cause oscillations near the end of the assembly. Hence, the constant compliance controller cannot ensure the well-performed stability and rapidity of assembly at the same time (Fig. 7 (a)).

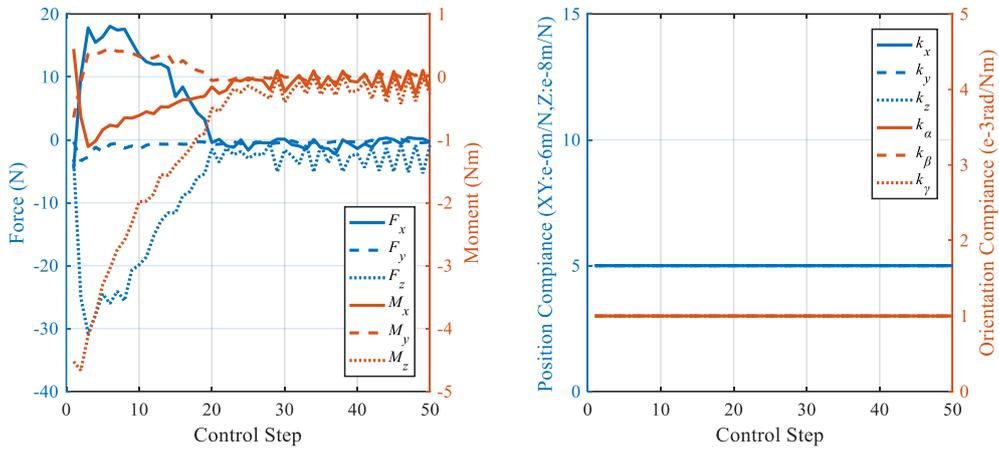

(a) Constant compliance controller.

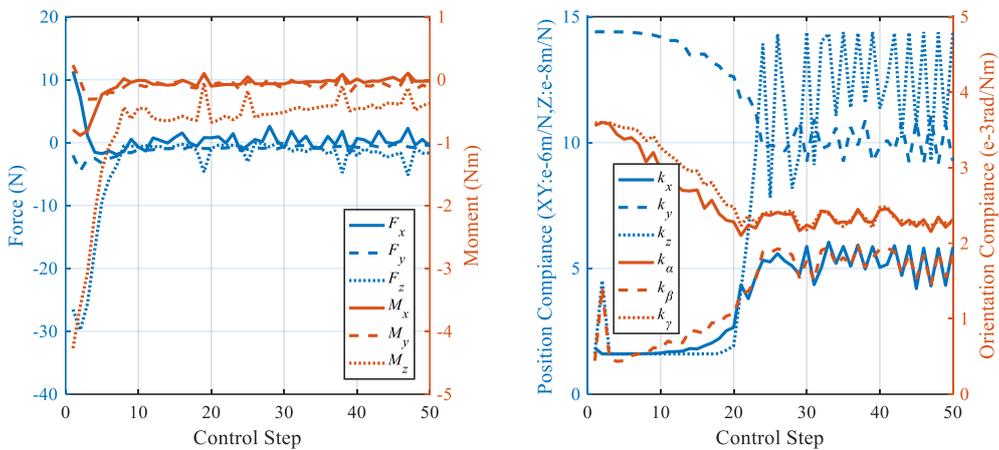

(b) Compliance controller combined with GCRL agent.

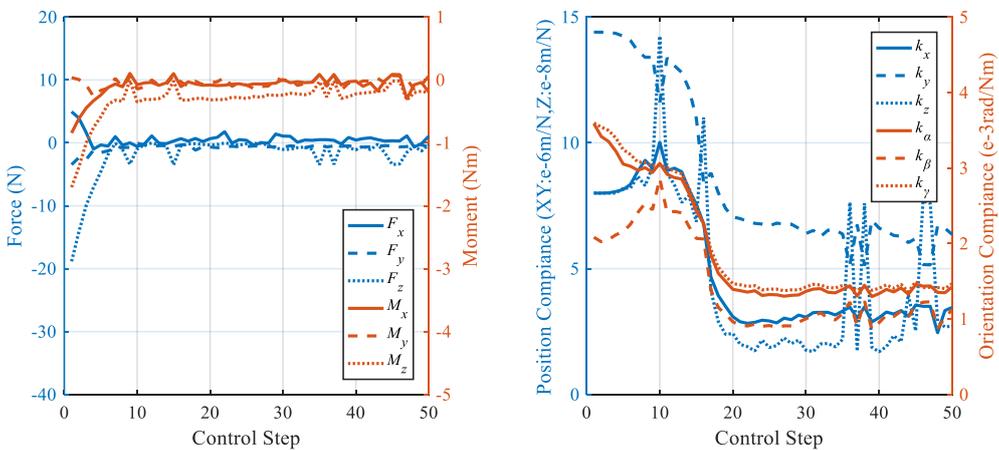

(c) Compliance controller combined with LCRL agent.

Fig. 7. The testing performance of controllers in simulations.

As shown in Fig. 7 (b) and (c), the adaptive compliance controller combined with an agent relates the compliance parameters with state. Therefore it is capable to improve the stability and rapidity of assembly together. As shown in Fig. 7 (b), because the irrelevant state components influence the selection of action, the adaptive compliance controller combined with global connection agent can only reduce assembly force/moment and improve assembly stability to some certain extent. As shown in Fig. 7 (c), LCRL method eliminates the effect of irrelevant state components on the selection of action. When the adaptive compliance controller combined with an LCRL agent is used to execute the assembly task, compliance parameters tend to be close to the upper boundary to maximize the control ability at the beginning of assembly. Then operation forces and moments start to decrease. With the decrease of operation force/moment, compliance parameters also decrease to stabilize the assembly process. In summary, LCRL method can obtain better stability and rapidity at the same time compared with GCRL method.

## 5. Experiments

### 5.1. Training process

Table III. Parameters in experiments.

| Variable | Value | Variable | Value |
|---|---|---|---|
| Axis length of pegs | 30 mm | Depth of holes | 30 mm |
| Side length of holes | 10 mm | Side length of pegs | 9.9 mm |
| Target insertion depth | 30 mm | Material of pegs and holes | ABS |
| Chamfer of pegs | 0.5 mm | Chamfer of holes | 0 mm |
| **Variable** | **Value** | | |
| Threshold $T$ | 0.1 | | |
| Target insertion step | 250 | | |
| Range of random pose | [± 0.4 mm, ± 0.4 mm, ± 0.4 mm, ± 1°, ± 1°, ± 1°] | | |
| Constant compliance | $\widehat{K}$ = diag(1e-3, 1e-3, 1e-5, 8e-4, 8e-4, 8e-4) | | |
| Action boundaries | $lb$ = [-1, -1, -1, -1, -1, -1], $ub$ = [2, 2, 2, 4, 4, 4] | | |
| Reward coefficients | $h_z$=1, $h_F$=0.1, $h_M$=1, $r_s$=2 | | |
| Learning rate of DDPG | 1e-3(actor), 1e-2(critic) | | |
| Soft update rate of DDPG | 1e-3 | | |

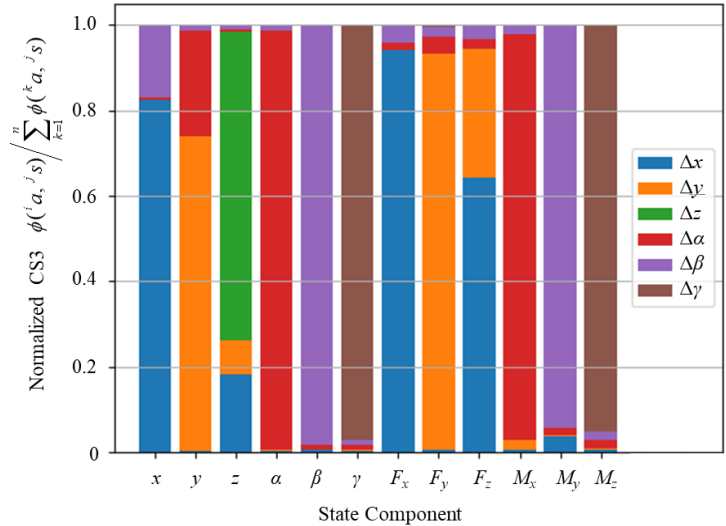

Fig. 8. Normalized CS3 for all action components and state components to construct the connection graph of the peg-in-hole assembly task in experiments.

In experiments, we use another set of equipment to perform the peg-in-hole assembly task (Fig. 3 (b)). The force/moment sensor is installed at the end of a parallel robot and the hole is fixed on the force/moment sensor. The parallel robot carries the hole to accomplish insertion. The force/moment sensor detects force/moment information caused by the contact between the peg and hole. The parameters of assembly objects and the controller are listed in Table III.

As shown in Fig. 8, we first obtain normalized CS3 for arbitrary *i* and *j* and construct a connection graph. Because modelling process in simulations is slightly different from the actual one in experiments, the connection graph in experiments is different with the simulation results. The change of $G_{22}$' is caused by coupling between the orientation of the robot and force/moment detected by the sensor, which exists in experiments but is not considered in the modelling of simulations.

$$G' = \begin{bmatrix} G_{11} & G_{12} \\ G_{21} & G_{22}' \end{bmatrix}, G_{22}' = \begin{bmatrix} 1 & 1 & 1 & 1 & 1 & 1 \\ 1 & 1 & 1 & 1 & 1 & 1 \\ 0 & 0 & 0 & 0 & 0 & 1 \end{bmatrix} \quad (23)$$

The local connection policy is set as

$$\begin{aligned}
\Delta x &= {}^1\pi(x, z, F_x, F_z, M_y) \\
\Delta y &= {}^2\pi(y, z, F_y, F_z, M_x) \\
\Delta z &= {}^3\pi(z) \\
\Delta \alpha &= {}^4\pi(y, \alpha, F_x, F_y, F_z, M_x, M_y, M_z) \\
\Delta \beta &= {}^5\pi(x, \beta, F_x, F_y, F_z, M_x, M_y, M_z) \\
\Delta \gamma &= {}^6\pi(\gamma, M_z)
\end{aligned} \quad (24)$$

Hence, the structure of LCRL agent is set according to (24). We use DDPG algorithm with LCRL method and GCRL method to train the controller separately. The reward curves of the training process are illustrated in Fig. 9. The results agree with simulation results. It takes LCRL method 51 episodes to reach -5 average reward, while it takes GCRL method 104 episodes. The final reward of LCRL method and GCRL method at 200 episode are -3.2 and -4.9. LCRL method has obvious positive significance in improving data-efficiency and final reward.

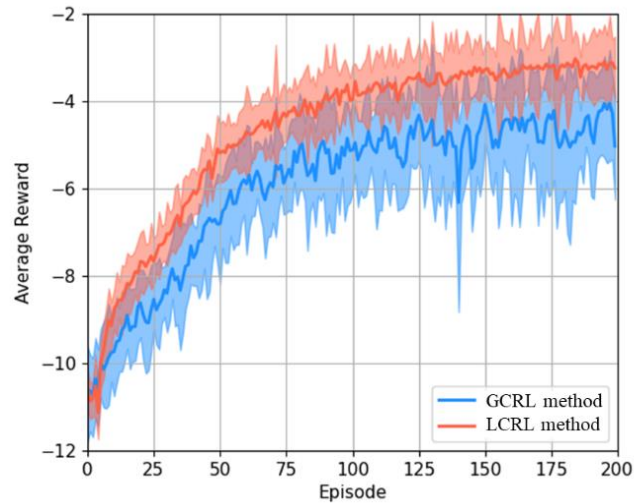

Fig. 9. The reward curves of training process in experiments.

## 5.2. Testing process

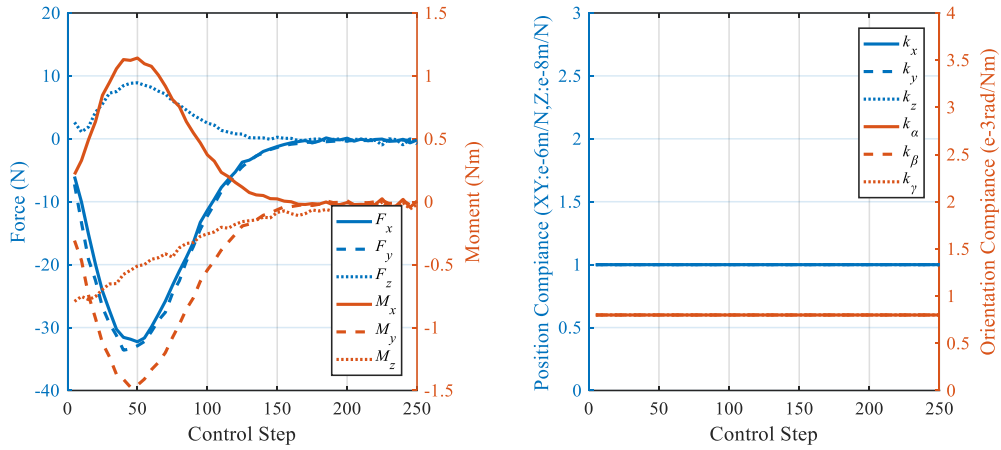

(a) Constant compliance controller.

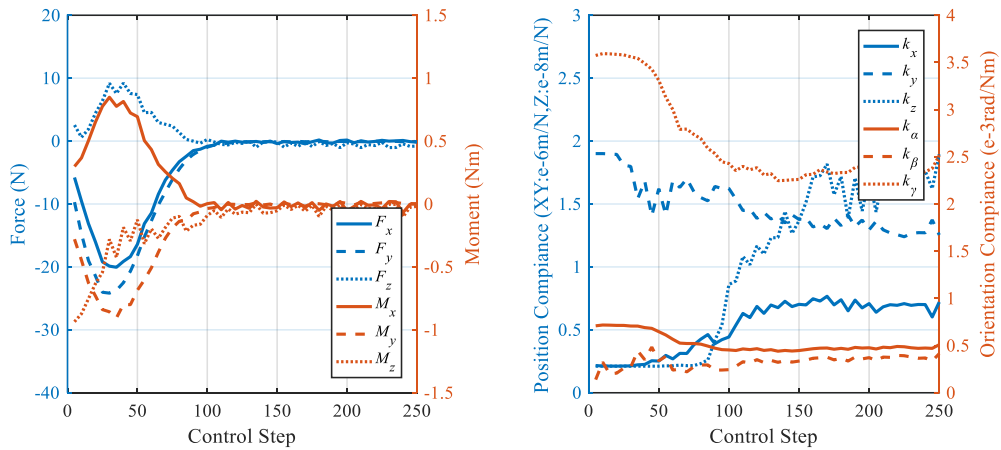

(b) Compliance controller combined with GCRL agent.

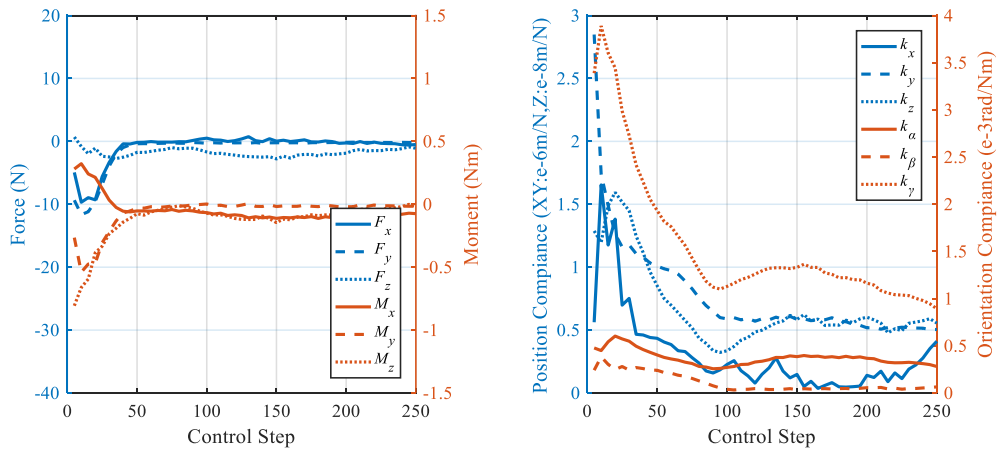

(c) Compliance controller combined with LCRL agent.

Fig. 10. The testing performance of controllers in experiments.

We test constant compliance controller, compliance controller combined with GCRL agent, and compliance controller combined with LCRL agent on the assembly task separately. As shown in Fig. 10, we compare the contact force/moment between the peg and hole when the initial pose error is set the same. The results are consistent with simulation results. The results demonstrate that LCRL method ensures smaller contact force and moment in the entire assembly process. Adaptive compliance controller correlates compliance parameters with state, which reflects the relationship between force/moment signals and the insertion depth as well as the complex force/moment-pose mapping in the assembly process, so that the rapidity and stability of assembly can be guaranteed simultaneously. Adaptive compliance controller combined with LCRL agent eliminates the influence of irrelevant state components on the selection of action, which provides more appropriate compliance parameters to the task, resulting in small and stable force/moment throughout the process.

## 6. Conclusions

The main contributions of this paper are summarized as follows. We first define CS3 to identify the effect of an action component on a state component and build a connection graph to describe the dependence of action on state. LCRL method is then proposed based on the connection graph to eliminate the influence of irrelevant state components on the selection of action, which improves the data-efficiency and final reward of the training process. Then we combine a compliance controller and an LCRL agent to accomplish robotic peg-in-hole assembly tasks. The combination of the LCRL agent provides the adaptive law for complex force/moment-pose mapping of assembly tasks, which improves stability and rapidity at the same time.

LCRL method is a general convergence optimization method of RL, which provides a new idea for RL to solve complex tasks. LCRL method is effective for the environment with local connection properties, of which the connection graph is sparse. The improvement effect of LCRL method is stronger in environments with sparser connection graphs. From the perspective of assembly tasks, this work is of substantial significance for precision assembly and complex shaped object assembly tasks. Applying LCRL method to more robotic manipulation fields is a significant and feasible future work.

## Appendix: Proof of Lemma 1

In a linear time-invariant system, state transition function can be written in recursive form. Hence,

$$|s_{t+1}' - s_{t+1}| = |Es_t' + Fa^{(1)} + Fa^{(2)} - Es_t + Fa^{(1)}| = |E(s_t' - s_t) + Fa^{(2)}|$$

When $t = 0$, we have

$$|s_1' - s_1| = |Es_0' + Fa^{(1)} + Fa^{(2)} - Es_0 + Fa^{(1)}| = |Fa^{(2)}|$$

Hence,

$$|s_{t+1}' - s_{t+1}| = |(E^t + E^{t-1} + ... + I)Fa^{(2)}|$$

Because

$$a^{(2)} \in {}^i\mathcal{A} = \{a \mid {}^k a = 0, k \neq i\}$$

One-step effect in $\phi({}^i a, {}^j s)$ at time $t$ in a linear time-invariant system can be written as

$$\mathbb{E}_{a^{(2)} \in {}^i\mathcal{A}, s_0' = s_0 \in \mathcal{S}} \left| {}^j s_{t+1}' - {}^j s_{t+1} \right| = C_{t,ij} \mathbb{E}_{a^{(2)} \in {}^i\mathcal{A}} \left| {}^i a \right|$$

where $C_{t,ij}$ is the element in row $i$ and column $j$ of matrix $C$. $C_{t,ij} \geq 0$.

$\phi(^{i}a, ^{j}s)$ is the sum of one-step effect. Hence, $\phi(^{i}a, ^{j}s)$ is derived as

$$\phi(^{i}a, ^{j}s) = \sum_{t=0}^{m-1} C_{t,ij} \mathbb{E}_{a^{(2)} \in ^{i}\mathcal{A}} \left| ^{i}a \right| = H_{ij} \mathbb{E}_{a^{(2)} \in ^{i}\mathcal{A}} \left| ^{i}a \right|$$

where $H_{ij}$ is the element in row $i$ and column $j$ of matrix $H$. $H_{ij} \geq 0$.

Because

$$\mathbb{E}_{a^{(2)} \in ^{i}\mathcal{A}} \left| ^{i}a \right| > 0$$

According to (6), $G_{ij} = 0$ if and only $H_{ij} = 0$, $G_{ij} = 1$ if and only $H_{ij} > 0$. *lemma* 1 is proved.

## Acknowledgements

This study was financially supported by the National Nature Science Foundation of China [Grant No. 51575306].